%% file: main.tex
\documentclass{article} 
\usepackage{mefomo_2023,times}

\input{math_commands.tex}

\usepackage{hyperref}
\usepackage{url}
\usepackage{booktabs, multirow} 

\usepackage{graphicx}
\graphicspath{ {./img} }
\usepackage{subfigure}

\usepackage{tabularx,ragged2e}

\usepackage{xcolor}

\newcommand\xspace{\mathcal{X}}
\newcommand\yspace{\mathcal{Y}}
\newcommand\zspace{\mathcal{Z}}
\newcommand\columns{\mathcal{C}}
\newcommand\num{\mathrm{num}}
\newcommand\cat{\mathrm{cat}}
\newcommand\FT{\mathrm{FT}}
\newcommand\AL{\mathrm{AL}}
\newcommand\RM{\mathrm{RM}}
\newcommand\enc{\mathrm{enc}}
\newcommand\PE{\mathrm{PE}}
\newcommand\dec{\mathrm{dec}}
\newcommand\concat{\mathrm{Concat}}
\newcommand\LN{\mathrm{LayerNorm}}
\newcommand\linear{\mathrm{Linear}}
\newcommand\mask{\mathtt{mask}}
\newcommand\pos{\mathrm{pos}}
\newcommand\proj{\mathrm{proj}}

\newcommand\RR{\mathbb{R}}

\title{TabRet: Pre-training Transformer-based\\ Tabular Models for Unseen Columns}

\iclrfinalcopy

\author{
Soma Onishi\thanks{Work done during internship at Preferred Networks, Inc.}\\
Meiji University\\
Kanagawa, Japan \\
\texttt{somaonishi4@gmail.com}
\And
Kenta Oono \& Kohei Hayashi \\
Preferred Networks, Inc. \\
Tokyo, Japan \\
\texttt{\{oono,hayasick\}@preferred.jp}
}

\begin{document}

\maketitle

\input{contents/abstract.tex}
\input{contents/introduction.tex}
\input{contents/related_work.tex}

\input{contents/method.tex}
\input{contents/experiments.tex}
\input{contents/conclusion.tex}



\bibliography{main}
\bibliographystyle{iclr2023_conference}

\appendix
\input{contents/network_architecture.tex}
\input{contents/experiment_detail.tex}
\input{contents/additional_experiment.tex}

\end{document}

%% file: math_commands.tex
\usepackage{amsmath,amsfonts,bm}

\def\eqref#1{equation~\ref{#1}}

\def\1{\bm{1}}

\def\vx{{\bm{x}}}

\DeclareMathAlphabet{\mathsfit}{\encodingdefault}{\sfdefault}{m}{sl}
\SetMathAlphabet{\mathsfit}{bold}{\encodingdefault}{\sfdefault}{bx}{n}

\newcommand{\R}{\mathbb{R}}



%% file: contents/abstract.tex
\begin{abstract}
We present \emph{TabRet}, a pre-trainable Transformer-based model for tabular data. TabRet is designed to work on a downstream task that contains columns not seen in pre-training. Unlike other methods, TabRet has an extra learning step before fine-tuning called \emph{retokenizing}, which calibrates feature embeddings based on the masked autoencoding loss. In experiments, we pre-trained TabRet with a large collection of public health surveys and fine-tuned it on classification tasks in healthcare, and TabRet achieved the best AUC performance on four datasets. In addition, an ablation study shows retokenizing and random shuffle augmentation of columns during pre-training contributed to performance gains. The code is available at \url{https://github.com/pfnet-research/tabret}.
\end{abstract}

%% file: contents/introduction.tex
\section{Introduction}

Transformer-based pre-trained models have been successfully applied to various domains such as text and images~\citep{bommasani2021on}. The Transformer-like architecture consists of two modules: a \emph{tokenizer}, which converts an input feature into a token embedding, and a \emph{mixer}, which repeatedly manipulates the tokens with attention and Feed-Forward Networks (FFN)~\citep{lin2021a,yu2021metaformer}. During pre-training, both modules are trained to learn representations that generalize to downstream tasks.

What has often been overlooked in the literature are scenarios where the input space change between pretext and downstream tasks.  A supervised problem on tabular data is a typical example, where rows or records represent data points and columns represent input features. Since the data scale is not as large as text and images, pre-trained models are expected to be beneficial~\citep{borisov2021deep}. A key challenge is that each table has a different set of columns, and it is difficult to know at the pre-training phase which columns will appear in the downstream task. We need to train the tokenizers from scratch for unseen columns with a small amount of data. Previous studies use text data such as a column name or a description to obtain the embeddings directly from language pre-trained models~\citep{wang2022transtab,hegselmann2022tabllm}, but we cannot do this when there is no such side information.

To address the above issue, we propose \emph{TabRet}, a pre-trainable Transformer network that can adapt to unseen columns in downstream tasks. First, TabRet is pre-trained based on the reconstruction loss with masking augmentation~\citep{devlin-etal-2019-bert,He_2022_CVPR}. Then, when unseen columns appear in a downstream task, their tokenizers are trained through masked modeling while freezing the mixer before fine-tuning, which we call \emph{retokenizing}. In experiments, we pre-trained TabRet with a table having more than two million rows and evaluated the performance on four tables containing $\sim 50$\% unseen columns. The results show TabRet outperformed baseline methods for all the datasets. Furthermore, the ablation study confirmed that retokenizing and random shuffling of columns within a batch further enhanced the pre-training effect.

\begin{figure}[tb]
\centering
\includegraphics[width=.95\textwidth]{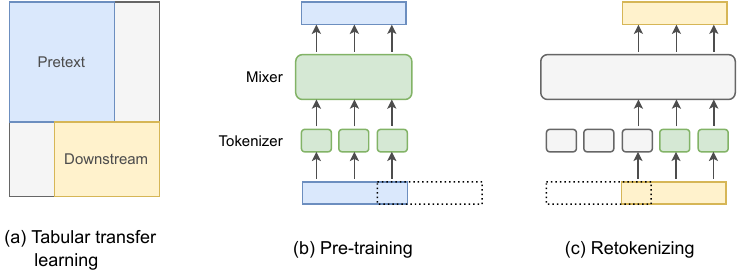}
\label{fig:concept}
\caption{\textbf{Problem setting and our approach.} (a) We consider transfer learning across tables that have different columns, where the pretext data is not available in the downstream task, and vice versa. (b) Our model consists of two modules: a tokenizer for each column and a mixer. We train the two modules based on the masked autoencoder loss in pre-training. (c) We introduce additional tokenizers for new columns and train them while freezing the old tokenizers and the mixer in a downstream task. }
\end{figure}

%% file: contents/related_work.tex
\section{Related Work}
Deep learning (DL)-based and tree-based methods are the two main streams of supervised learning for tabular data.
Which one works better depends on the task, as they capture different aspects of the input characteristics~\citep{grinsztajn2022why,SHWARTZZIV202284}.
However, DL-based methods are primarily adapted to transfer learning because they are easier to pre-train than tree-based methods.
Among others, self-supervised learning with Transformers is the most common pre-training approach of DL-based methods (see, e.g.,~\citet{badaro:hal-03877085} for the survey).

Despite the popularity, most Transformer-based methods have not adapted to the case where pre-training and fine-tuning tasks have different column sets.
To the best of our knowledge, TransTab~\citep{wang2022transtab}, TabLLM~\citep{hegselmann2022tabllm}, and LIFT~\citep{dinh2022lift} are the few exceptions that can train on tabular datasets with different column sets.
TransTab creates column-by-column representations of each row from the column name and value, allowing pre-trained models to be applied to unseen columns.
TabLLM and LIFT convert a row into a sequence of tokens and feed them to a pre-trained language model~\citep{sanh2021multitask, gpt-j}.
They assumed there were semantic correspondences of column descriptions between different columns.
In practice, however, there are many cases where column descriptions are not informative (e.g., random alphabet) or do not exist. 
Since TabRet does not explicitly use the column description information, it can transfer to a different column set in such a situation.

%% file: contents/method.tex
\section{Method}\label{sec:method}

\paragraph{Problem Formulation.}
In pre-training, suppose we have a table consisting of a finite set of columns $\columns$. Let $\xspace_c$ be the space where column $c\in\columns$ takes its value. A row $\vx$ is then defined on the product space $\xspace(\columns)=\prod_{c\in\columns}\xspace_c$. Suppose a downstream task is defined as a supervised task on an input-output pair $(\vx', y)$. Here we consider the case where the input $\vx'\in\xspace(\columns')$ is a row  of another column set $\columns'\neq\columns$ (Figure~\ref{fig:concept} a). As an example, consider healthcare records where $\columns=\{\mathtt{age}, \mathtt{gender}, \mathtt{weight}\}$ are given as a column set in pre-training and $\columns'=\{\mathtt{gender}, \mathtt{BMI} \}$ are in fine-tuning. In this case, the column $\mathtt{gender}$ appears in common, but other columns $\{\mathtt{age}, \mathtt{weight}, \mathtt{BMI}\}$ appear only in either pre-training or fine-tuning. We assume that the pre-training table is inaccessible during fine-tuning. 

\paragraph{Model Structure.} 
Given a set of columns $\columns$, we define a tokenizer $t_c: \xspace_c \to \mathcal{E}$ for each column $c\in \columns$ as a function that converts the column value $x_c\in\xspace_c$ into an embedding vector $e_c\in\mathcal{E}$. The tokenizers then map an entire row $\vx$ to a set of embeddings $\{e_c \mid c\in\columns\}$, which are passed to a transformer-based encoder $h: \mathcal{E}^{\columns} \to \zspace$ to produce a latent representation $z\in\zspace$. We also use a decoder $d: \zspace \to \xspace(\columns)$ to reconstruct the input in pre-training/retokenizing and a head $p: \zspace \to \yspace$ to predict the target variable in fine-tuning. 
Note that part of the decoder and the head have the same network structure, and their parameters are partially shared. More details about the network architecture are described in Appendix~\ref{sec:model}.

\paragraph{Pre-training with Shuffle Augmentation.} \label{sec:shuffle-augmentation}
We train the tokenizers of the columns $\columns$, the encoder, and the decoder by following the approach of the masked autoencoder~\citep{He_2022_CVPR}; that is, we randomly select the columns with a masking ratio and replace the corresponding embeddings with a special embedding called a mask token. We then reconstruct the values of the masked columns and compute the loss to update the parameters. We empirically find that the masking ratio = 0.7 works well for pre-training and 0.5 for retokenizing. We use these values throughout the experiments unless otherwise mentioned.

During pre-training, we apply the shuffle augmentation as follows. Let $\vx_c = (x_{1c}, \dots, x_{nc})\in\xspace_c^n$ be a batch of column $c$ of size $n$ and $\mathrm{perm}(\cdot)$ be a random permutation. Suppose $\tilde{\columns} \subseteq\columns$ is a set of columns chosen uniformly randomly based on a shuffle ratio. Then the shuffle augmentation replaces $\vx_c$ with $\mathrm{perm}(\vx_c)$ for $c \in \tilde{\columns}$. We set the shuffle ratio to 0.1.

\paragraph{Retokenizing and Fine-tuning.} 
In a downstream task, we first add initialized tokenizers for the newly appearing columns. Part of the decoder is also initialized to match the fine-tuning table. Then we train them by masked autoencoding. During this time, we do not update the parameters of the old tokenizers, encoder, and decoder (Figure~\ref{fig:concept} c). Afterward, the head is added to the backbone network (tokenizer + encoder) and fine-tuned to predict the target variable. 

The primary motivation for retokenizing is to efficiently train the new tokenizers with a relatively small number of data points. Although we can train them in a supervised manner, the model is easily overfitted because, in addition to the tokenizers, the head must be trained with a single target signal. In contrast, the masked autoencoding loss used in retokenizing provides more signals to reconstruct than the target variable, which is expected to induce better token representations. 

%% file: contents/experiments.tex
\section{Experiments}\label{sec:experiments}
\paragraph{Datasets.}
As pre-training data, we preprocessed behavioral risk factor surveillance system (BRFSS), a collection of public health surveys in the US, and created a single table consisting of 2.03 million rows and 74 columns. As downstream tasks, we selected four classification datasets in the healthcare domain from Kaggle: Diabetes, HDHI, PKIHD, and Stroke.
Each dataset has about 50\% overlap with BRFSS columns. However, some of these overlapping columns have different representations of their values. For example, the age column in BRFSS is categorical, but is represented as continuous in the downstream datasets, such as Stroke. We pre-processed the column representations of the downstream datasets to adjust BRFSS columns.
We used 20\% of the data as a test set, 100 data points from 80\% for fine-tuning (and retokenizing), and the remaining data as a validation set. 
The dataset specifications are described in Appendix~\ref{sec:datasets}.

\paragraph{Baselines.}

We used two groups of baselines: supervised methods trained only on the downstream tasks and self-supervised methods pre-trained on BRFSS. As supervised methods, we compared logistic regression (LR), XGBoost, CatBoost, MLP, and Feature Tokenizer Transformer~\citep{gorishniy2021revisiting} (FTTrans). Self-supervised methods were SCARF~\citep{bahri2022scarf} and TransTab~\citep{wang2022transtab}. The hyperparameters, such as learning rate, were optimized by Optuna~\citep{10.1145/3292500.3330701} with the validation set for each method.
Details of the baseline methods are described in Appendix~\ref{sec:baselines}.

\paragraph{Results.}

Table~\ref{tab:ood} shows that TabRet outperformed the baselines for all the datasets. Among the pre-trained models, TabRet consistently achieved the best. In addition to absolute performance, TabRet had relatively little variability (i.e., small variance) in the results, which is one of the desirable properties of pre-training.

We also ablated TabRet. Since the default masking ratios mentioned in Section~\ref{sec:method} were determined based on the performance with all the learning options (pre-training, shuffle augmentation, retokenizing), we also optimized them as hyperparameters here for a fair comparison. The results (Table~\ref{tab:ablation}) show a positive trend of pre-training. Although shuffle augmentation and retokenizing sometimes worked negatively, they provide additional gains on average. 

To make a fair comparison, we evaluated if the shuffle augmentation is effective for TransTab. However, the performance did not consistently improve (refer to Appendix\ref{sec:additional_exp_transtab}).

\begin{table}[tb]
 \caption{\textbf{Test AUC performance.} The methods from LR to FTTrans were trained only on fine-tuning data (i.e., no pre-training). Each cell reports the mean and standard deviation over 20 random seeds. The best scores are underlined, and those with statistical significance at a significance level of 0.05 (Welch’s t-test) are in bold. 
 }
 \label{tab:ood}
 \centering
  \begin{tabular}{rcccc}
   Methods & Diabetes & HDHI & PKIHD & Stroke\\
   \midrule
   LR & $75.10\pm 3.55$ & $75.55\pm 3.43$ & $76.91\pm 2.52$ & $74.29\pm 6.08$ \\
   XGBoost & $79.52\pm 0.79$ & $80.29\pm 1.25$ & $79.74\pm 0.93$ & $69.02\pm 9.63$ \\
   CatBoost & $77.83\pm 1.40$ & $77.65\pm 1.82$ & $76.50\pm 1.79$ & $76.14\pm 3.46$ \\
   MLP & $78.20\pm 1.02$ & $79.39\pm 1.09$ & $77.51\pm 1.68$ & $76.27\pm 5.92$ \\
   FTTrans & $79.11\pm 1.07$ & $78.96\pm 1.30$ & $76.45\pm 2.38$ & $76.48\pm 4.92$ \\
   \midrule
   SCARF & $78.43\pm 1.35$ & $80.36\pm 1.26$ & $81.01\pm 0.94$ & $76.74\pm 5.04$ \\
   TransTab & $78.30\pm 1.18$ & $78.77\pm 1.34$ & $78.56\pm 1.58$ & $75.00\pm 4.80$ \\
   TabRet & \underline{$79.94\pm 1.03$} & $\mathbf{81.65\pm 1.60}$ & $\mathbf{82.70\pm 0.79}$ & $\mathbf{80.73\pm 3.83}$ \\
  \end{tabular}
\end{table}

\begin{table}[tb]
\caption{\textbf{Ablation study.} \emph{Supervised} indicates TabRet that was trained only with fine-tuning data. \emph{Supervised} row reports AUC scores and the remaining rows report the performance gains averaged over 10 random seeds.}
\label{tab:ablation}
\centering
\begin{tabular}{lllllr}
    & Diabetes & HDHI & PKIHD & Stroke & Ave. Gain\\
    \midrule
    Supervised       & $78.71$ & $78.42$ & $75.47$ & $74.29$ & N/A\\
    \midrule
    + Pre-training   & $+0.12$ & $+2.06$ & $+4.98$ & $+6.12$ & $+3.07$\\
    + Shuffle aug.   & $+0.90$ & $+2.86$ & $+4.99$ & $+5.16$ & $+3.21$ \\
    + Retokenizing & $+0.64$ & $+2.65$ & $+6.89$ & $+5.30$ & $+3.87$
\end{tabular}
\end{table}

%% file: contents/conclusion.tex
\section{Discussion}

Tabular data have been notorious for transfer learning due to the difference in column sets. We addressed the problem and presented the transformer network with two additional steps --- random shuffling in pre-training and retokenizing before fine-tuning --- excelled in its potential as a pre-trained model. We hope our results will open a new research direction of pre-trained models on tabular data or other data domains where the input space can be changed. 

The current limitation is that pre-training does not always provide performance gain, especially when the domain of a downstream task is irrelevant to the pretext data. We evaluated the pre-trained model used in the experiments on several datasets having no column overlap. The preliminary result was that the performance was worse than the supervised methods. It may be because the generative data process is entirely different, and there is no transferable knowledge in the task pair. Another possibility is that there was transferable knowledge, but our framework failed to capture it. Further investigation of this topic would be promising as future work. 

%% file: contents/network_architecture.tex
\section{Model}\label{sec:model}

\begin{figure*}[htbp]
\centering
\subfigure[Pre-training phase]{
\includegraphics[width=0.45\textwidth]{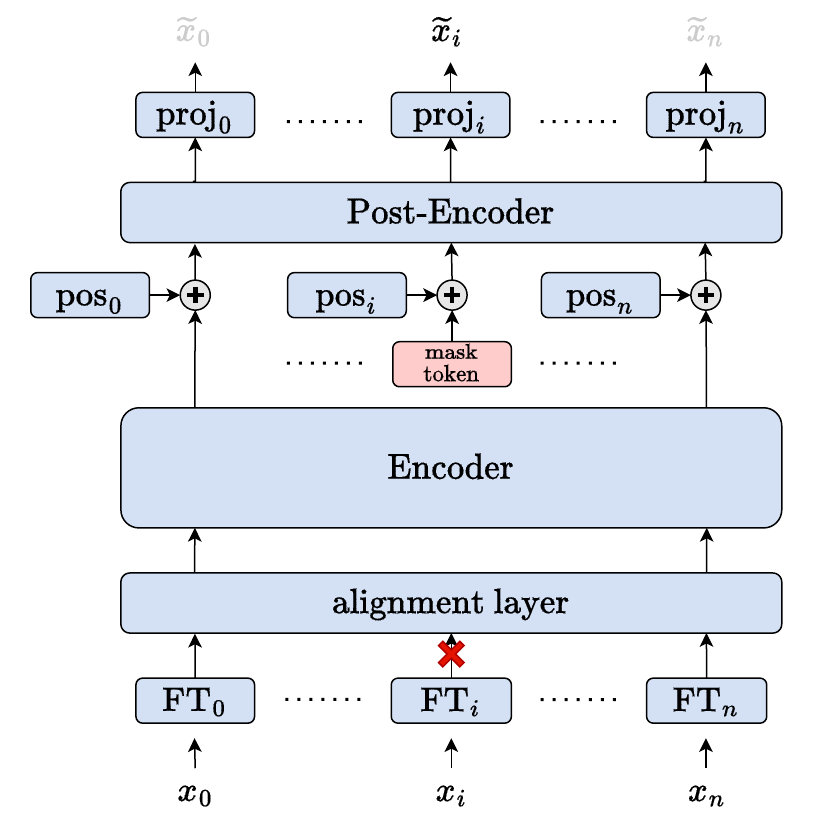}
\label{fig:model_pre}
}
\subfigure[Retokenizing phase]{
\includegraphics[width=0.45\textwidth]{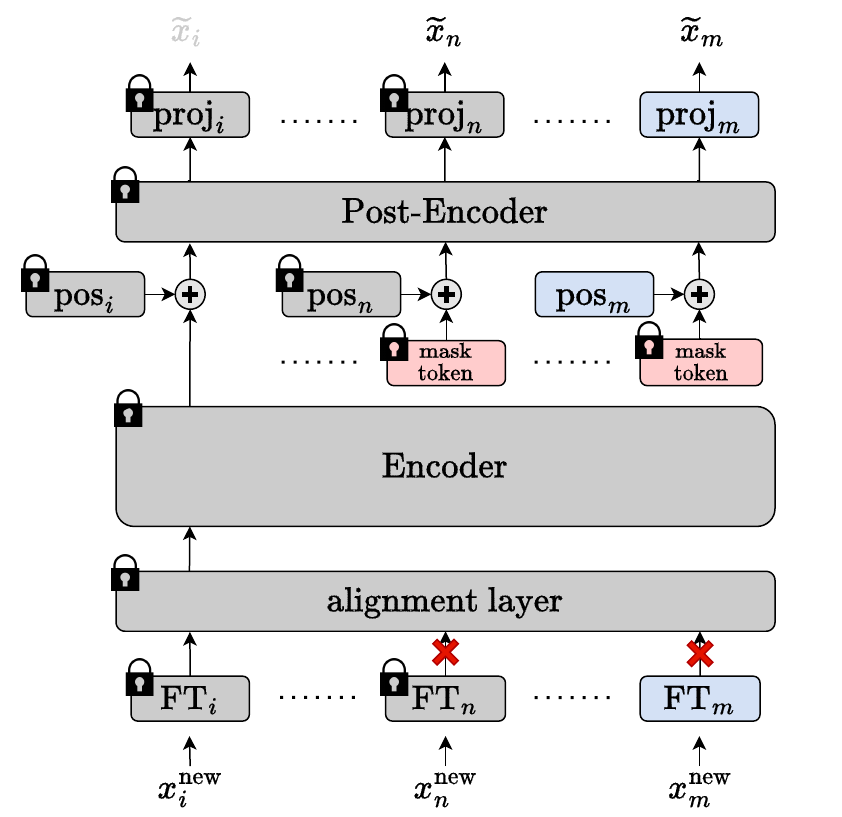}
\label{fig:model_column_align}
}
\subfigure[Fine-tuning phase]{
\includegraphics[width=0.45\textwidth]{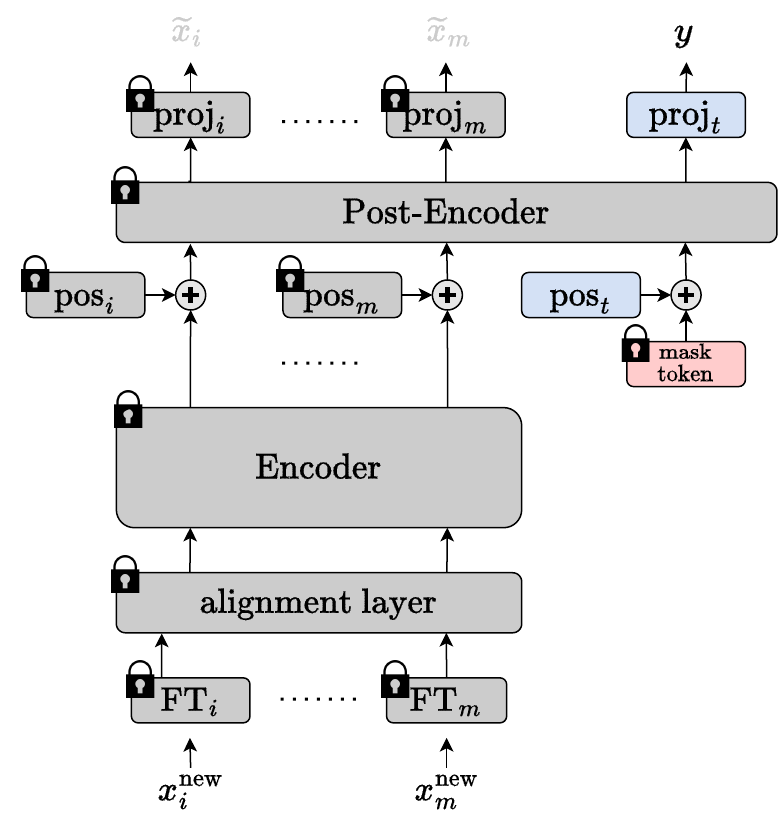}
\label{fig:model_fine}
}
\caption{Schematic view of our model and training phases. (a): Pre-training phase trains all modules by reconstructing masked features. (b) Retokenizing phase trains the modules corresponding to newly added columns. In this figure, the $0$-th to $(i-1)$-th columns are removed, and the $(n+1)$-th to $m$-th columns are added. Feature Tokenizer $\FT_j$, Positional Embedding $\pos_j$, and Projector $\proj_j$ are trained for $j=n+1,\ldots, m$. (c): Fine-tuning phase freezes all parameters trained in the pre-training and retokenizing phases and trains the positional embedding and Projector for the target variable ($\pos_t$ and $\proj_t$ in the figure, respectively). }
\label{fig:model_abs}
\end{figure*}

\subsection{Architecture}\label{sec:architecture}

\subsubsection{Overview}

This section describes the architecture of our proposed model TabRet.
TabRet consists of Feature Tokenizer, Alignment Layer, Random Masking, Encoder, Post-Encoder, and Projector.
Feature Tokenizer converts each feature into a token. Alignment Layer normalizes the set of tokens and adjusts the tokens' dimensionality.
Random Masking masks some tokens and feeds the remaining tokens to Encoder.
Masked tokens are replaced with the mask token $[\mask]$ before Post-Encoder.
Finally, Projector sends back tokens to the input domains.
Figure~\ref{fig:model_abs} shows the schematic view of our model.

\subsubsection{Feature Tokenizer}

Feature Tokenizer module converts an input $\vx$ to an embedding $T^{\FT}$, whose dimensions will be determined later.
We assume that each column is either numerical or categorical. 
Since the order of the feature does not matter, as we see in this section, we write the input as $\vx=(x_1^{\num}, \ldots, x_{k^{\num}}^{\num}, x_1^{\cat}, \ldots, x_{k^{\cat}}^{\cat})$.
Here, $k^{\num}$ and $k^{\cat}$ are the number of numerical and categorical features, respectively.
$x_j^{\num}\in \R$ is the $j$-th numerical feature and $x_j^{\cat} \in [C_j]$ is the $j$-th categorical feature, where $C_j$ is the number of categories of $j$-th categorical feature.

Feature Tokenizer embeds numerical features using the weight matrix $W^{\num}\in \R^{k\times d_{\FT}}$ and the bias vector $b^{\num}\in \R^d$ as follows:
\[
    T_j^{\num} = x_j^{\num}W_j^{\num} + b_j^{\num} \in \RR^{d_{\FT}}.
\]
For categorical features, Feature Tokenizer converts the input $x_j^{\cat}$ to a one-hot vector $e_j^{\cat} \in \{0, 1\}^{C_{j}}$ and embeds it using the lookup table $W^{\cat} \in \RR^{C_j \times d_{\FT}}$ and bias $b^{\cat} \in \RR^{d_{\FT}}$:
\[
    T_j^{\cat} = e_j^\top W_j^{\cat} + b_j^{\cat} \in \mathbb{R}^{d_{\FT}}.
\]
$W^{\num}$, $b^{\num}$, $W^{\cat}$, $b^{\cat}$ are learnable parameters.
Then, embeddings are concatenated for further processing:
\[
    T^{\FT} = \concat(T_1^{\mathrm{num}}, \ldots, T_{k^{\mathrm{num}}}^{\mathrm{num}}, T_1^{\mathrm{cat}}, \ldots, T_{k^{\mathrm{cat}}}^{\mathrm{cat}}) \in \RR^{k\times d_{\FT}},
\]
where $k = k^{\num} + k^{\cat}$.

\subsubsection{Alignment Layer}

We employed different dimensions for Feature Tokenizer and Encoder for the model's flexibility.
Alignment Layer changes the token dimensions to adjust Encoder using a linear layer.
Alignment Layer also normalizes the scale of tokens using Layer Normalization~\cite{ba2016layer}:
\[
    T^{\AL} = \linear(\LN(T^{\FT})) \in \RR^{k\times d_{\AL}},
\]
where $d_{\AL}$ is the dimension of tokens that Alignment Layer outputs.

\subsubsection{Random Masking}

Random Masking behaves differently depending on training phases (see Appendices~\ref{sec:pretext-task} and~\ref{sec:downstream-task} for the definition of the phases.)
In the pre-training and retokenizing phases, Random Masking masks some tokens randomly.
More specifically, we set the mask ratio $\alpha$, chose $m'=\lfloor \alpha k \rfloor$ tokens uniformly randomly from the set of tokens for each data point and dropped the chosen tokens.
If $m'=0$, we overrode the value of $m'$ by $1$. That is, Random Masking removes one token uniformly randomly.
Consequently, the feature size becomes from $T^{\AL}\in \RR^{k\times d_{\AL}}$ to $T^{\RM}\in \RR^{m\times d_{\AL}}$, where $m = k-m'$ is the number of tokens that are not dropped.
In the fine-tuning phase, Random Masking does nothing, that is, $T^{\RM} = T^{\AL}$.

\subsubsection{Encoder}

Encoder is an $N$-layer Transformer.
We use the Pre-Norm variant, which is reportedly easy for optimization~\cite{1906.01787}.
In addition, we add one Layer Normalization after the final Transformer block.
Mathematically, the computation of Encoder can be described as follows:
\begin{align*}
    T_0 &= T^{\RM},\\
    T_i &= F_i(T_{i-1}) \quad \text{$i = 1, \ldots N$},\\
    T^{\enc} &= \LN(T_N),
\end{align*}
where $F_i$ is the $i$-th Transformer block.
The output $T^{\enc}$ of Post-Encoder is the $m$-tuple of $d_{\enc}$-dimensional tokens: $T^{\enc} = (T^{\enc}_1, \ldots, T^{\enc}_m)\in \RR^{m\times d_{\enc}}$.

\subsubsection{Post-Encoder}

Post-Encoder first applies a linear layer to project embeddings to the Post-Encoder's input dimension $d_{\PE}$.
Then, a mask token $[\mask]$, which is a $\RR^{d_{\PE}}$-dimensional learnable vector, is inserted into each position where Random Masking removes the encoder token.
Additionally, Post-Encoder adds learnable positional embedding $[\pos] = ([\pos]_1, \ldots, [\pos]_k)\in \RR^{k\times d_{\PE}}$ to the embeddings, which are expected to learn column-specific information.
Finally, the embeddings are transformed by a one-layer Transformer block $F$.
In summary, the architecture of Post-Encoder is as follows:
\begin{align*}
    T_1 &= \linear(T^{\enc}),\\
    T_2 &= \mathrm{AddMaskToken}(T_1, \mathrm{[\mask]}),\\
    T_3 &= T_2 + [\pos],\\
    T_4 &= F(T_3),\\
    T^{\PE} &= \LN(T_4).
\end{align*}
Similarly to Encoder, the output $T^{\PE}$ of Post-Encoder is the $k$-tuple of $d_{\PE}$-dimensional tokens: $T^{\PE} \in \RR^{k\times d_{\PE}}$.

\subsubsection{Projector}

Projector sends back the tokens to column feature spaces using linear layers.
We prepare one projector per column since each column has different scales and dimension.
Specifically, let $T^{\dec} = (T^{\num}_1, \ldots, T^{\num}_{k_\num}, T^{\cat}_1, \ldots, T^{\cat}_{k_\cat})\in \RR^{k\times d_{\PE}}$ be the Post-Encoder's output.
Then, the output of the $j$-th projector is as follows:
\[
    \hat{\vx}^{\ast}_j = \linear(T_j^{\mathrm{\ast}}),
\]
where $\ast \in \{\num, \cat\}$.
The dimensionality of $\hat{x}^{\num}_j$ is $1$ for numerical features for all $j=1, \ldots, k^{\num}$, and $\hat{x}^{\cat}_j$ is $C_j$ for the $j$-th categorical feature for $j=1, \ldots, k^{\cat}$.

We can think of our model as an encoder-decoder model by interpreting the pair of Post-Encoder and Projector as a decoder.
While the original masked autoencoder in~\citet{He_2022_CVPR} uses its decoder only in the pretext task and removes it in the downstream task, our model uses the decoder trained on the pretext task in the downstream tasks.
This architectural difference comes from the modality of data.
In~\citet{He_2022_CVPR}, whose modality of interest is images, models have to solve qualitatively different problems as the pretext and downstream tasks -- for example, image inpainting for the pretext task object classification for the downstream task.
On the other hand, in tabular learning settings, both pretext and downstream tasks are supervised learning tasks on columns.
We expect the decoder is more likely to learn the knowledge beneficial for the downstream task in the fine-tuning phase.
Therefore, we design our model to reuse the same decoder in the fine-tuning phases.

\subsection{Training}

\subsubsection{Pretext Task}\label{sec:pretext-task}

The pretext task consists of a single phase, namely, the pre-training phase (Figure~\ref{fig:model_pre}).

\paragraph{Shuffle Augmentation}

\begin{figure}[tb]
    \centering
    \includegraphics[width=0.3\textwidth]{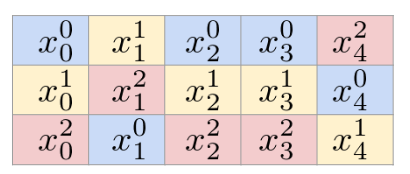}
    \caption{Schematic view of shuffle augmentation. Shuffle augmentation applies column-wise permutation within a minibatch to a randomly chosen subset of columns. This figure shows the shuffle ratio $\beta=0.4$, and the first and fourth columns out of five are shuffled.}
    \label{fig:column_shuffle}
\end{figure}

In the pre-training phase, we apply shuffle augmentation (Figure~\ref{fig:column_shuffle}) to the input minibatches.
We set the shuffle ratio $\beta \in (0, 1)$.
Given a minibatch $\vx$, we choose $\ell = \lfloor \beta k \rfloor$ features uniformly randomly.
For each chosen column, we permute the features in the minibatch as described in Section~\ref{sec:shuffle-augmentation}.

\subsubsection{Downstream Task}\label{sec:downstream-task}
We considered two types of downstream tasks.
The first one was the independently-and-identically-distributed (IID) transfer learning (Appendix~\ref{sec:additional_experiment}), where pretext and downstream tasks have the same column sets.
In the IID transfer learning setting, the downstream task has a single phase, namely, the fine-tuning phase (Figure~\ref{fig:model_fine}). The task is equivalent to the usual supervised learning task in the IID case.
The other type was the out-of-distribution (OOD) transfer learning (Section~\ref{sec:experiments}), where column sets of pretext and downstream tasks could be different.
The downstream task in the OOD transfer learning additionally has the retokenizing phase (Figure~\ref{fig:model_column_align}) before the fine-tuning phase.

\paragraph{Retokenizing Phase}

TabRet has Feature Tokenizer, Positional Embedding, and Projector for each column, as explained in Appendix~\ref{sec:architecture}.
Therefore, we needed to learn these modules for columns unseen in the pretext task.
In the retokenizing phase, we freeze all parameters except these modules for columns unseen in the pretext task.
We employ the same masked modeling training as the pretext task.
We treat the unseen columns as masked and feed the frozen mask tokens $[\mask]$ as the input to Post-Encoder corresponding to the columns.

\paragraph{Fine-tuning Phase}
The target value $y$ of the downstream task is treated as a masked entry in the newly added column during the fine-tuning phase. All parameters learned during the pre-training and retokenizing phases are frozen. The Positional Embedding and Projector are then trained for the target value column, similarly to the retokenizing phase. By doing so, this approach significantly reduces the number of learning parameters, which in turn reduces the required sample size of the training dataset for the downstream task. Furthermore, we note that our model does not use the [\texttt{cls}] token, unlike the approach by~\citet{He_2022_CVPR}. 

\subsubsection{Loss Function}

For pre-training and retokenizing phases, we define the loss value $L(\vx)$ of the data point $\vx$ as the sum of losses for those features that are masked by Random Masking and are not shuffled by the shuffle augmentation:
\begin{equation*}
    L(\vx) = \sum_{j = 1}^{k^{\num}} \ell^{\num}(x_j^{\num}, \hat{x}_j^{\num}) + \sum_{j = 1}^{k^{\cat}} \ell^{\cat}(x_j^{\cat}, \hat{x}_j^{\cat}).
\end{equation*}
For fine-tuning phase, we compute the loss value $L(\vx, y)$ of a data point $(\vx, y)$ as the loss for its target value:
\begin{equation*}
    L(\vx, y) =
    \begin{cases}
        \ell^{\num}(y, \hat{y}) & \text{if $y$ is numerical,}\\
        \ell^{\cat}(y, \hat{y}) & \text{if $y$ is categorical,}
    \end{cases}
\end{equation*}
where $\hat{y}$ is the output of Projector corresponding to the target feature when the input is $\vx$.
The loss function $\ell^{\num}$ for numerical features is the mean squared loss, and the loss function $\ell^{\cat}$ for categorical features is the cross-entropy loss.
The training objective is the sum of the loss values for all instances in the training dataset.

%% file: contents/experiment_detail.tex
\section{Details of Experiment Settings}

\subsection{Datasets}\label{sec:datasets}
\begin{table}[t]
 \caption{Dataset specifications. All datasets are binary classification tasks. Cat: the number of categorical features. Num.: the number of numerical features. Positive: the ratio of data points with the positive target. Overlap: the ratio of columns that exist in the pre-training dataset BRFSS.}
 \label{tab:dataset}
 \centering
  \begin{tabular}{lrrrrr}
    Name & Data Points & Cat. & Num. & Positive & Overlap \\
   \midrule
   Diabetes & 253,680 & 20 & 1 & 0.139 & 0.524 \\
   HDHI & 253,680 & 20 & 1 & 0.094 & 0.524 \\
   PKIHD  & 319,795 & 15 & 2 & 0.086 & 0.529 \\
   Stroke & 4,909 & 2 & 8 & 0.049 & 0.5 \\
  \end{tabular}\label{sec:dataset-specification}
\end{table}

We used 4 datasets in OOD experiment.
Table~\ref{sec:dataset-specification} summarizes the specifications of the datasets and Table~\ref{sec:dataset-source} lists their sources.

\subsubsection{Pre-training Datasets}

\begin{table*}[t]
 \caption{Splitting of BRFSS dataset. The numbers represent the sample sizes. The fine-tuning and test data were used in the experiments in Appendix~\ref{sec:additional_experiment}.}
 \label{tab:diabetesl}
 \centering
  \begin{tabular}{cccc}
   \multicolumn{4}{c}{\textbf{All}} \\
   \multicolumn{4}{c}{2,038,772} \\ 
   \midrule
   \multicolumn{3}{c}{\textbf{Train}} & \textbf{Test} \\
   \multicolumn{3}{c}{1,631,018} & 497,754 \\
   \cmidrule(lr){1-3}
   \textbf{Pre-training} & \textbf{Validation} & \textbf{Fine-tuning} & \\
   1,453,235 & 163,102 & 100 &\\
  \end{tabular}
\end{table*}

We used telephone survey datasets obtained from Behavioral Risk Factor Surveillance System (BRFSS)\footnote{\url{https://www.cdc.gov/brfss}} as a pre-training dataset.
These datasets collected state-specific risk behaviors related to chronic diseases, injuries, and preventable infectious diseases of adults in the United States.
We combined the datasets from 2011--2015 and removed missing values by deleting rows and columns by the following steps: 1. deleted columns with more than 10\% missing values; 2. deleted rows with missing values.
We call the resulting dataset \textbf{BRFSS}.
We split the BRFSS as shown in Table~\ref{tab:diabetesl} and used the pre-train and validation datasets. 

\subsubsection{Fine-tuning Datasets}

We prepared four datasets for the OOD Transfer Learning:
\begin{itemize}
    \item \textbf{Diabetes}: The dataset made from the BRFSS dataset of 2015 to predict whether a subject had diabetes. 
    \item \textbf{HDHI} (Heart Disease Health Indicator): The dataset made from the BRFSS dataset of 2015 to predict whether the subject has heart disease.
    \item \textbf{PKIHD} (Personal Key Indicator of Heart Disease): Similarly to HDHI, the task is to predict whether the patient has heart disease, but was made from the BRFSS dataset of 2020.
    \item \textbf{Stroke}: The dataset recording clinical events. The task is to predict whether a subject is likely to get a stroke using 10 features: gender, age, hypertension, heart disease, marriage, work type, residence type, glucose, BMI, and smoking status.
\end{itemize}
These datasets were divided into 80\% training dataset and 20\% test dataset.
Of the training data, 100 samples were separated as a fine-tuning dataset.

 In some columns, BRFSS and the above four data are named differently but have the same meaning. Therefore, to create overlap, the column names of the above four data were changed to match the column names in BRFSS. In addition, we performed the same feature engineering performed in BRFSS, such as categorizing age.

\subsubsection{Data Preprocessing}

For all methods except the tree-based methods, numerical features were transformed using the Quantile Transformation from the scikit-learn library~\citep{JMLR:v12:pedregosa11a}, and categorical features were transformed using the Ordinal Encoder. For the tree-based methods, only categorical features were transformed using the Ordinal Encoder.

\subsection{Baselines}\label{sec:baselines}

\begin{table}[t]
\caption{Hyperparameters of Transformer layer in TransTab. The default value of n\_layer is 2. FFN stands for Feed-Forward Network and is often used in the Transformer layer.}
\label{tab:hypara_transformer_transtab}
 \centering
 \begin{tabular}{l|ccc}
n\_layer & 2 & 6 \\
\midrule
Hidden\_size & 128 & 384 \\
Num\_attention\_head & \multicolumn{2}{c}{8} \\
Hidden\_dropout\_prob & \multicolumn{2}{c}{0.0} \\
FFN\_dim & 128 & 512\\
Activation & \multicolumn{2}{c}{ReLU} \\
\end{tabular}
\end{table}

We compared our model with the following baselines:

\begin{itemize}
  \item \textbf{Logistic Regression} (LR). We implemented it using \texttt{LogisticRegression} module of the scikit-learn package
  \item \textbf{Multi-layer Perceptron} (MLP). A simple deep learning model. We implemented using PyTorch~\citep{NEURIPS2019_bdbca288}. 
  \item \textbf{Gradient Boosting Decision Tree}: One of the most standard tree-based algorithms. We used two implementations in the experiments, \textbf{XGBoost}~\citep{Chen:2016:XST:2939672.2939785} and \textbf{CatBoost}~\citep{NEURIPS2018_14491b75}.   
  \item \textbf{FT-Transformer}~\citep{gorishniy2021revisiting}. One of the standard Transformer models for tabular data, on which our model is based. We implemented using \texttt{rtdl} package\footnote{\url{https://github.com/Yura52/rtdl}}.
  \item \textbf{SCARF}~\citep{bahri2022scarf} \textbf{with FT-Transformer}. Self-supervised learning for tabular data using contrastive learning. Although~\citet{bahri2022scarf} employed a multi-layer perceptron as an encoder, we substituted it with FT-Transformer to remove the effect of the choice of backbones. In the pre-training phase, the class token in the Encoder's output was fed to Projector, and computed the infoNCE loss using the output of Projector. In the fine-tuning phase, the class token of the Encoder's output was fed to the classification head to solve the downstream task. We set the number of Encoder's blocks to 6. Details of Encoder are shown in Table~\ref{tab:hypara_transformer}.
  \item \textbf{TransTab}~\citep{wang2022transtab}. A transformer-based model that supports transfer learning across different column sets. TransTab assumes that column names have semantic meanings, which may not be the case in the BRFSS dataset (see Section~\ref{sec:experiments} and Appendix~\ref{sec:datasets} for details of dataset characteristics.) For a fair comparison, we used the same Feature Tokenizer as our model for the tokenization of features. We set the number of Encoder’s blocks to 6. Details of Encoder are shown in Table~\ref{tab:hypara_transformer_transtab}.
\end{itemize}

We applied only supervised learning using the fine-tuning dataset to those methods that did not support transfer learning, namely LR, MLP, GBDT, and FT-Transformer.
For SCARF and TransTab, we used part of the authors' implementation and modified the model as described above.
Even for those models, we implemented the training part of the models by ourselves.
For the methods that require pre-training, including our model, we trained the pre-trained models with multi-node distributed learning using PyTorch's \texttt{DistributedDataParallel}.

\subsection{Implementation of TabRet}
\begin{table}[t]
\caption{Hyperparameters of Transformer layer. FFN stands for Feed-Forward Network and is often used in the Transformer layer.}
\label{tab:hypara_transformer}
 \centering
 \begin{tabular}{l|cccccc}
n\_blocks & 1 & 2 & 3 & 4 & 5 & 6 \\
\midrule
Token size & 96 & 128 & 192 & 256 & 320 & 384 \\
Head count & \multicolumn{6}{c}{8} \\
Activation \& FFN size factor & \multicolumn{6}{c}{(ReGLU, 4/3)} \\
Attention dropout & \multicolumn{6}{c}{0.1} \\
FFN dropout & 0.0 & 0.05 & 0.1 & 0.15 & 0.2 & 0.25 \\
Residual dropout & \multicolumn{6}{c}{0.0} \\
Initialization & \multicolumn{6}{c}{Kaiming} \\
\end{tabular}
\end{table}

In all experiments, unless otherwise specified, We set the number of Encoder blocks to 6. Details of each hyperparameter are shown in Table~\ref{tab:hypara_transformer}. We changed only the FFN dropout of Encoder to 0.1. The output dimension of the Feature Tokenizer and the input/output dimension of the Alignment Layer were aligned with the input dimension of the Encoder: $d_\FT = d_\AL = d_\enc$.

\subsection{Training and Hyperparameter Optimization}
\label{sec:training_tuning}

We performed hyperparameter optimization with Optuna~\citep{10.1145/3292500.3330701} for all methods except Logistic Regression, where the number of optimization trials was 500 trials for the tree-based method and 100 trials for the DL-based method. And we set the early stopping patience to 20 for all methods except Logistic Regression.

\subsubsection{Non-DL Methods}
\paragraph{Logistic Regression (LR).} We used the default settings, except that the maximum number of iterations was set to 1000.
\paragraph{XGBoost.} The hyperparameter space for Optuna is shown in Table~\ref{tab:xgb}.
\paragraph{CatBoost.} The hyperparameter space for Optuna is shown in Table~\ref{tab:cat}.

\begin{table}[ht]
 \caption{XGBoost hyperparameter space, the same hyperparameter space searched by \citet{grinsztajn2022why}. We used defaults for the other hyperparameters.}
 \label{tab:xgb}
 \centering
  \begin{tabular}{ll}
    Parameter & Distribution\\
    \midrule
    max\_depth & UniformInt[1, 11]\\
    n\_estimators & UniformInt[100, 6000, 200]\\
    min\_child\_weight & UniformInt[1, 1e2]\\
    subsample & Uniform[0.5, 1.0]\\
    learning\_rate & LogUniform[1e-5, 0.7]\\
    colsample\_bylevel & Uniform[0.5, 1.0]\\
    colsample\_bytree & Uniform[0.5, 1.0]\\
    gamma & LogUniform[1e-8, 7]\\
    lambda & LogUniform[1, 4]\\
    alpha & LogUniform[1e-8, 1e2]\\
  \end{tabular}
\end{table}

\begin{table}[ht]
 \caption{CatBoost hyperparameter space. We used defaults for the other hyperparameters.}
 \label{tab:cat}
 \centering
  \begin{tabular}{ll}
    Parameter & Distribution\\
    \midrule
    max\_depth & UniformInt[3, 10]\\
    learning\_rate & LogUniform[1e-5, 1]\\
    bagging\_temperature & Uniform[0, 1]\\
    l2\_leaf\_reg & LogUniform[1, 10]\\
    leaf\_estimation\_iterations & UniformInt[1, 10]\\
  \end{tabular}
\end{table}

\subsubsection{DL Methods}
\begin{table}[tb]
    \centering
    \caption{Detail about training for DL methods.}
    \label{tab:training_configs}
    \begin{tabular}{l|rr}
    Config & Pre-training & Fine-tuning \\
    \midrule
    optimizer & \multicolumn{2}{c}{AdamW} \\
    weight decay & \multicolumn{2}{c}{1e-5}\\
    optimizer momentum & \multicolumn{2}{c}{$\beta_1, \beta_2 = 0.9, 0.99$}\\
    learning rate schedule & \multicolumn{2}{c}{cosine decay~\citep{loshchilov2017sgdr}}\\
    \midrule
    epochs & 1000 & 200\\
    base learning rate & 1.5e-5 & Searched by Optuna \\
    batch size & 4096 & 32\\
    warmup epochs~\citep{https://doi.org/10.48550/arxiv.1706.02677} & 40 & 5\\
    \end{tabular}
\end{table}

Table~\ref{tab:training_configs} shows the details of training for DL methods. For pre-training and fine-tuning, we use the linear $lr$ scaling rule~\citep{https://doi.org/10.48550/arxiv.1706.02677} for both pre-training and fine-tuning: $lr = base\_lr \times batchsize / 256$.

\paragraph{Multi-layer Perceptron (MLP).} The hyperparameter space for Optuna is shown in Table~\ref{tab:mlp}. Note that the sizes of the first and last layers were tuned and set separately, while the size of the in-between layers is the same for all of them.
\paragraph{FT-Transfomrer.} The hyperparameter space for Optuna is shown in Table~\ref{tab:fttrans}. Other parameters, such as token size, are determined from the corresponding parameters in Table~\ref{tab:hypara_transformer} according to the selected n\_blocks.
\paragraph{SCARF.} We corrupted features independently using the Bernoulli distribution. For all datasets, the probability of corruption was fixed at 0.6. The Feature Tokenizer and Encoder parameters learned during pre-training were frozen for all fine-tuning phases, as it was confirmed that this improved the fine-tuning accuracy. The hyperparameter space for Optuna is shown in Table~\ref{tab:scarf}.
\paragraph{TransTab.} We used Self-VPCL proposed by \citet{wang2022transtab} and pre-trained with overlap ratio = 0.1 and num\_partition = \{2, 3, 4\}. The hyperparameter space for Optuna is shown in Table~\ref{tab:transtab}.
\paragraph{TabRet.} For all datasets, the mask ratio for pre-training was fixed at 0.7 and the mask ratio for retokenizing was fixed at 0.5.The hyperparameter space for Optuna is shown in Table~\ref{tab:ours}.

\begin{table}[ht]
 \caption{MLP hyperparameter space.}
 \label{tab:mlp}
 \centering
  \begin{tabular}{ll}
    Parameter & Distribution\\
    \midrule
    layers & UniformInt[1, 8]\\
    layer\_size & UniformInt[1, 512]\\
    dropout & Uniform[0, 0.5]\\
    category\_embedding\_size & UniformInt[64, 512]\\
    base\_learning\_rate & LogUniform[1e-4, 1e-1]\\
  \end{tabular}
\end{table}

\begin{table}[ht]
 \caption{FT-Transformer hyperparameter space.}
 \label{tab:fttrans}
 \centering
  \begin{tabular}{ll}
    Parameter & Distribution\\
    \midrule
    n\_blocks & UniformInt[1, 6]\\
    base\_learning\_rate & LogUniform[1e-4, 1e-1]\\
  \end{tabular}
\end{table}

\begin{table}[ht]
 \caption{SCARF hyperparameter space.}
 \label{tab:scarf}
 \centering
  \begin{tabular}{ll}
    Parameter & Distribution\\
    \midrule
    base\_learning\_rate & LogUniform[1e-4, 1e-1]\\
  \end{tabular}
\end{table}

\begin{table}[ht]
 \caption{TransTab hyperparameter space.}
 \label{tab:transtab}
 \centering
  \begin{tabular}{ll}
    Parameter & Distribution\\
    \midrule
    num\_partitions & UniformInt[2, 4]\\
    base\_learning\_rate & LogUniform[1e-4, 1e-1]\\
  \end{tabular}
\end{table}

\begin{table}[ht]
 \caption{TabRet hyperparameter space.}
 \label{tab:ours}
 \centering
  \begin{tabular}{ll}
    Parameter & Distribution\\
    \midrule
    base\_learning\_rate (Retokenizing) & LogUniform[1e-4, 1e-1]\\
    base\_learning\_rate & LogUniform[1e-4, 1e-1]\\
  \end{tabular}
\end{table}

%% file: contents/additional_experiment.tex
\section{Additional Experiments}
\subsection{TransTab with shuffle augmentation}
\label{sec:additional_exp_transtab}
For a fair comparison, the test AUC performance for fine-tuning in TransTab applied shuffle augmentation during pre-training has been presented in Table~\ref{tab:transtab_shuffle}. It was observed that TransTab applied shuffle augmentation during pre-training did not result in a consistent performance improvement.

\begin{table}[t]
    \centering
    \caption{Test AUC performance. Comparison of Transtab with and without shuffle augmentation during pre-training. We set the shuffle ratio to 0.1, similar to the shuffle ratio in TabRet.}
    \label{tab:transtab_shuffle}
    \begin{tabular}{lcccc}
    Method & Diabetes & HDHI & PKIHD & Stroke \\
    \midrule
    TransTab & $78.30\pm 1.18$ & $78.77\pm 1.34$ & $78.56\pm 1.58$ & $75.00\pm 4.80$  \\
    TransTab w/ shuffle aug. & $77.21\pm 1.30$ & $77.75\pm 1.71$ & $79.36\pm 1.50$ & $76.26\pm 4.82$
    \end{tabular}
\end{table}

\subsection{IID Experiment}
\label{sec:additional_experiment}

We investigate how TabRet is competitive with the existing self-supervised tabular models in the same column setting.

\subsubsection{Datasets}

We used four popular benchmarks listed in Table~\ref{tab:ssl_data}, in addition to BRFSS.

\begin{table}[t]
    \centering
    \caption{Self-Supervised datasets.}
    \label{tab:ssl_data}
    \begin{tabular}{lrrrr}
    Name & Data Points & Cat. & Num. & Positive \\
    \midrule
    BRFSS (2011--2015) & 2,038,772 & 64 & 10 & 0.127 \\
    Adult (AD) & 32,561 & 8 & 6 & 0.241 \\
    Bank Marketing (BM) & 45,211 & 10 & 6 & 0.117 \\
    HTRU2 (HR) & 17,898 & 0 & 8 & 0.092 \\
    Online Shoppers (OS) & 12,330 & 7 & 10 & 0.155 \\
    \end{tabular}
\end{table}

 Each of the newly added datasets was split into 4 subset, each of which is for pre-training (\texttt{pre}), fine-tuning (\texttt{fine}), validation (\texttt{val}), and test (\texttt{test}). The ratio of these subsets is as follows:

\begin{enumerate}
    \item \texttt{all} $\rightarrow$ \texttt{train}$_0$ : \texttt{test} = 0.8 : 0.2
    \item \texttt{train}$_0$ $\rightarrow$ \texttt{train}$_1$ : \texttt{val} = 0.9 : 0.1
    \item \texttt{train}$_1$ $\rightarrow$ \texttt{pre} : \texttt{fine}$_0$ = 0.8 : 0.2
    \item \texttt{fine}$_0$ $\rightarrow$ \texttt{fine} : \texttt{none} = 100 : Remaining
\end{enumerate}

The splitting of BRFSS is shown in Table~\ref{tab:diabetesl}.

\subsubsection{Baseline}

We compared FTTrans, SCARF, and TransTab. Except for BRFSS, we decreased the batch size for pre-training to 512. We used 3 transformer blocks for SCARF and TabRet (see Table~\ref{tab:hypara_transformer}; the number of Post-Encoder blocks is 1) and 2 for TransTab as the default value of \citep{wang2022transtab}. Other settings were the same as described in Appendix~\ref{sec:training_tuning}.

\subsubsection{Result}

\begin{table*}[t]
 \caption{Test AUC performance in IID setting. The number of random seeds is 10. }
 \label{tab:ssl_result}
 \centering
  \begin{tabular}{lccccc}
   Method & BRFSS & AD & BM & HR & OS \\
   \midrule
   FTTrans & $77.77\pm 1.18$ & $88.11\pm 0.46$ & \underline{$86.92\pm 0.70$} & $96.82\pm 0.47$ & \underline{$89.89\pm 0.68$}\\
   \midrule
   SCARF & $75.36\pm 2.27$ & $88.00\pm 0.29$ & $74.92\pm 2.38$ & $96.34\pm 0.41$ & $80.19\pm 3.75$\\
   TransTab & $77.41\pm 2.11$ & \underline{$88.32\pm 0.45$} & $85.65\pm 1.35$ & $96.74\pm 0.49$ & $89.85\pm 0.83$\\
   TabRet & $\mathbf{79.67\pm 1.19}$ & $88.08\pm 0.39$ & $76.93\pm 1.21$ & \underline{$96.90\pm 0.39$} & $84.96\pm 2.21$\\
  \end{tabular}
\end{table*}

Table~\ref{tab:ssl_result} shows the results. TabRet outperformed the baselines on BRFSS, but the performance was not stable for other datasets. In contrast to the pre-trained models, FTTrans achieved competitive performance for all the datasets. The results suggest the size of the datasets was not large enough to surpass the supervised method.

\begin{table}[tb]
 \caption{Dataset source.}
 \label{tab:dataset-url}
 \centering
  \begin{tabularx}{\textwidth}{lX}
    Dataset Name & URL \\
   \midrule
    BRFSS & \tiny\url{https://www.kaggle.com/datasets/cdc/behavioral-risk-factor-surveillance-system} \\
    Diabetes & \tiny\url{https://www.kaggle.com/datasets/alexteboul/diabetes-health-indicators-dataset} \\
    HDHI & \tiny\url{https://www.kaggle.com/datasets/alexteboul/heart-disease-health-indicators-dataset} \\
    PKIHD & \tiny\url{https://www.kaggle.com/datasets/kamilpytlak/personal-key-indicators-of-heart-disease} \\
    Stroke & \tiny\url{https://www.kaggle.com/datasets/fedesoriano/stroke-prediction-dataset}\\
    Adult & \tiny\url{https://archive.ics.uci.edu/ml/machine-learning-databases/adult/}\\
    Bank Marketing & \tiny\url{https://archive.ics.uci.edu/ml/datasets/bank+marketing}\\
    HTRU2 & \tiny\url{https://archive.ics.uci.edu/ml/datasets/HTRU2}\\
    Online Shoppers & \tiny\url{https://archive.ics.uci.edu/ml/datasets/Online+Shoppers+Purchasing+Intention+Dataset}\\
  \end{tabularx}\label{sec:dataset-source}
\end{table}